%% file: ms.tex
\documentclass[10pt,twocolumn,letterpaper]{article}

\usepackage{iccv}
\usepackage{times}
\usepackage{epsfig}
\usepackage{graphicx}
\usepackage{amsmath}
\usepackage{amssymb}

\usepackage{caption}
\usepackage{subcaption}
\usepackage{comment}

\usepackage[pagebackref=true,breaklinks=true,letterpaper=true,colorlinks,bookmarks=false]{hyperref}

\iccvfinalcopy %

\def\iccvPaperID{3901} %

\ificcvfinal\fi

\begin{document}

\iftrue
	\definecolor{mncolor}{RGB}{255,50,00}
	\newcommand\MATTHIAS[1] {\textbf{\textcolor{mncolor}{MN: #1}}}
	\definecolor{jtcolor}{RGB}{0,0,255}
	\newcommand\JT[1] {\emph{\textcolor{jtcolor}{JT: #1}}}
	\definecolor{todocolor}{RGB}{255,0,00}
	\newcommand\TODO[1] {\emph{\textcolor{todocolor}{TODO: #1}}}
	\newcommand\rev[1] {\emph{\textcolor{todocolor}{#1}}}
\else 
	\newcommand\MATTHIAS[1] {}
	\newcommand\JT[1] {}
	\newcommand\TODO[1] {}
\fi 

\title{Dynamic Surface Function Networks for Clothed Human Bodies}

\author{
Andrei Burov$^1$~~~
Matthias Nie{\ss}ner$^1$~~~
Justus Thies$^{1,2}$~~~
\vspace{0.2cm} \\ 
$^1$Technical University of Munich~~~
$^2$Max Planck Institute for Intelligent Systems, Tübingen
\vspace{0.2cm}
}

\ificcvfinal\thispagestyle{empty}\fi

\twocolumn[{
	\renewcommand\twocolumn[1][]{#1}%
	\maketitle
	\begin{center}
		\includegraphics[width=1.0\textwidth]{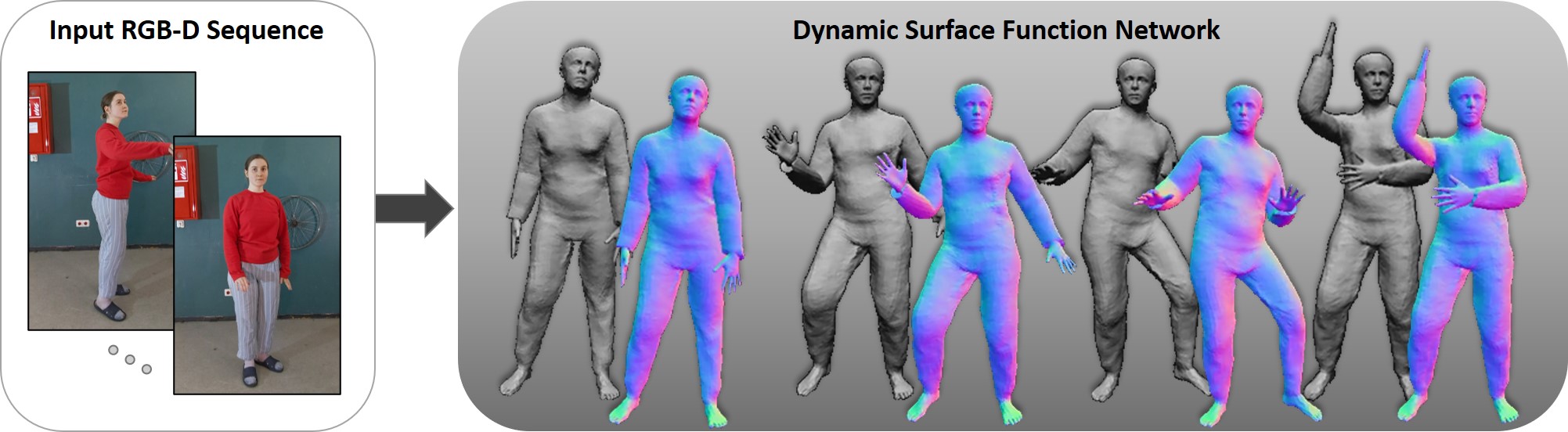}
		\captionof{figure}{We introduce dynamic surface function networks which are trained on a short RGB-D input video sequence of a clothed human.
		Our continuous surface representation allows the modeling of clothes as well as the dynamic pose dependent deformations.
		We parametrize this representation by the kinematic model of SMPL~\cite{SMPL2015}, which facilitates full pose control at inference time (e.g., through joint rotations). Here, we show an animation sequence taken from the MOSH~\cite{mosh,amass} dataset.
		}
		\label{fig:teaser}
	\end{center}
}]

\input{0_abstract}
\input{1_introduction}
\input{2_relatedwork}

\input{3_method}
\input{4_results}

\input{5_conclusion}

\vfill

\input{7_acknowledgments}

{\small
\bibliographystyle{ieee_fullname}
\bibliography{egbib}
}

\clearpage
\begin{appendix}
\input{6_supplement}
\end{appendix}

\end{document}

%% file: 0_abstract.tex
\begin{abstract}

We present a novel method for temporal coherent reconstruction and tracking of clothed humans.
Given a monocular RGB-D sequence, we learn a person-specific body model which is based on a dynamic surface function network.
To this end, we explicitly model the surface of the person using a multi-layer perceptron (MLP) which is embedded into the canonical space of the SMPL body model.
With classical forward rendering, the represented surface can be rasterized using the topology of a template mesh.
For each surface point of the template mesh, the MLP is evaluated to predict the actual surface location.
To handle pose-dependent deformations, the MLP is conditioned on the SMPL pose parameters.
We show that this surface representation as well as the pose parameters can be learned in a self-supervised fashion using the principle of analysis-by-synthesis and differentiable rasterization.
As a result, we are able to reconstruct a temporally coherent mesh sequence from the input data.
The underlying surface representation can be used to synthesize new animations of the reconstructed person including pose-dependent deformations.

\end{abstract}

%% file: 1_introduction.tex
\section{Introduction}

Digital capture of human bodies is a rapidly growing research area in computer vision and computer graphics.
There are many high impact applications, in particular, in the field of telepresencing and man-machine interaction that rely on the reconstruction of the surface as well as the motion of a person. 
For instance, for telepresencing in virtual reality (VR) or augmented reality (AR), the ultimate goal is to photo-realistically re-render the people who want to interact with each other.
A key challenge is to capture and reproduce the natural motions, including the dynamically changing surface (especially clothing).
In our work, we present a novel representation for the surface of a clothed human called \textit{dynamic surface function networks} which captures pose-dependent surface deformations such as wrinkles of the clothing.
In contrast to recent works on implicit representations~\cite{chibane20ifnet,bhatnagar2020ipnet,Huang:ARCH:2020}, our representation explicitly models the surface of the human.
Both representations have their advantages and disadvantages.
We use an explicit surface representation to leverage fast forward rendering (exploiting the rasterization units of a GPU), and global surface correspondences between different frames.
Specifically, our dynamic surface function network is attached to the surface of the SMPL~\cite{SMPL2015} body model which gives us access to the kinematic chain, as well as to a topology for rendering.
In particular, our model represents a continuous offset surface that can be evaluated at arbitrary points of the SMPL surface (also within triangles).
However, note that our representation is agnostic to the underlying parametric model and can also be used for other dynamically changing surfaces, e.g., human faces.
The dynamic surface function network is trained in a person-specific fashion.
To this end, we assume a short sequence (few seconds) of the person moving in front of a single consumer-level RGB-D camera.
Note that we do not assume any specific motion sequences (e.g., standing in T-pose or alike).
We jointly optimize the surface representation network as well as the pose parameters of the underlying SMPL body model.
This global optimization strategy allows us to fuse all captured data into a consistent surface representation.
This person-specific surface model can be animated explicitly using the joint control handles of the SMPL model, thus, allowing pose transfer (see Fig.~\ref{fig:teaser}).

\medskip
\noindent
To summarize, we present a method that allows reconstruction of a person-specific controllable body model based on a monocular input sequence captured by a commodity RGB-D sensor.
Our key contributions are:
\begin{itemize}
    \item an explicit surface representation network which is able to model the pose-dependent deformations of the surface, such as wrinkles of the clothing.
    \item a global analysis-by-synthesis formulation that allows for the joint optimization of the pose and the surface over the entire sequence, leading to a temporally consistent tracking of the person in the input sequence.
\end{itemize}

%% file: 2_relatedwork.tex
\section{Related Work}

Our work is based on an \textit{explicit surface representation}.
In the literature, explicit surface representations (especially, triangle meshes) are the most prominent representations for human faces and bodies~\cite{SMPL2015,SMPL-X:2019,zollhoefer2018facestar}.
However, implicit representations are also used to represent the surface of a body including the clothing~\cite{bhatnagar2020ipnet,chibane20ifnet,deng2019neural,Huang:ARCH:2020,pifuSHNMKL19,saito2020pifuhd}.
Implicit surface representations have the advantage of not needing special care for handling topological changes (as would be required for an explicit representation like a mesh).
On the other hand, implicit representations do not provide explicit correspondences over a time series.
For instance, methods such as Occupancy Flow~\cite{Niemeyer2019ICCV} predict scene flow, but are limited by the sequence length.
Methods like \cite{genova2019sif,genova2020ldif,tretschk2020patchnets,bozic2021neuraldeformationgraphs} reconstruct implicit function of an object's geometry in a patch-based manner and empirically show consistency of patches across a sequence.
PIFu~\cite{pifuSHNMKL19} and PIFuHD~\cite{saito2020pifuhd} are able to reconstruct a pixel-aligned implicit function only based on RGB input images.
They leverage a learned prior trained on a synthetic dataset of humans.
Similarly, IF-Nets~\cite{chibane20ifnet} learn a prior to reconstruct an implicit function of a human, based on pointcloud or depth-map inputs.
IF-Nets have been extended in a follow-up work called IP-Nets~\cite{bhatnagar2020ipnet} to fit an explicit surface to the reconstructed implicit surface (SMPL model + displacements).

Aside from these per-frame reconstruction methods, there exist methods that incrementally fuse observations into a discretized implicit function (volumetric SDF grid)~\cite{curless1996volumetric}.
The seminal work of DynamicFusion by Newcombe et al.~\cite{Newcombe_2015_CVPR} is able to reconstruct dynamically changing objects only based on a depth sequence.
Follow-up works~\cite{innmann2016volume,Guo2017,dou20153d,guo2017monofvv} added additional color constraints or dense SDF alignment \cite{slavcheva2017killingfusion,slavcheva2018sobolevfusion}.
BodyFusion~\cite{BodyFusion} and DoubleFusion~\cite{DoubleFusionPAMI_2019} added a deformation prior using a human skeleton.
In contrast to these methods, our approach reconstructs a controllable mesh including pose-dependent deformations, allowing us to animate the reconstructed mesh.
Note that these fusion methods work on a frame-to-frame tracking scheme.

Our method is an optimization-based approach that jointly optimizes the surface over the entire input sequence and does not require a learned prior as \cite{bhatnagar2020ipnet,chibane20ifnet,pifuSHNMKL19,saito2020pifuhd}.
The global optimization scheme of our method is closely related to MonoClothCap~\cite{xiang2020monoclothcap}.
MonoClothCap gets an RGB sequence as input to track, and reconstructs a clothed human using an explicit surface representation.
In order to handle clothing, they rely on a PCA model and shading-based refinement of the mesh.
Another technique to model the surface of clothing explicitly has been shown in CAPE~\cite{ma20autoenclother}.
They learn a variational auto-encoder to represent offsets for clothing relative to the SMPL model based on high quality multi-view reconstructions using ClothCap~\cite{ponsmollSIGGRAPH17clothcap}.
This learned prior allows them to reconstruct clothed humans from single RGB inputs only, including pose-dependent deformations.
Alldieck et al.~\cite{alldieck2019tex2shape} reconstruct a detailed human body shape based on a single RGB input leveraging a learned aligned image-to-image translation technique to regress texture maps for the geometry and color of the model.
In contrast, our approach is not based on a learned prior and optimizes the surface based on the observations from an RGB-D camera.
Our dynamic surface function network represents the pose-dependent geometry and is not restricted to a static surface~\cite{Bogo:ICCV:2015,Weiss:ICCV:11,Zhang2014,helten2013}.
Methods that estimate the body shape under cloth~\cite{shapeundercloth:CVPR17,wuhrer2013,stoll2010,layered_model,Pons-Moll_MRFIJCV} can be used as an initialization of our method.
Our approach works with a consumer-grade RGB-D camera, and does not require a multi-camera setup to learn or reconstruct a person-specific template in advance~\cite{habermann2019TOG,habermann20deepcap}.
We use RGB-D data to explore the representation power of our surface network, but we see the potential of our representation to be used for video-based reconstruction similar to \cite{alldieck2018video,alldieck20183DV,alldieck19cvpr}.
Note that our method is not designed for real-time use such as LiveCap~\cite{habermann2019TOG}, since our focus lies on the global reconstruction of a controllable mesh with a pose-dependent surface.
Concurrent work analyzes other representations of a human surface that are based on points~\cite{ma2021pop}, local surface patches~\cite{scale}, or implicit representations for shape and motions~\cite{palafox2021npm,leap,Wang2021CVPR,scanimate,MetaAvatar:arXiv:21,liu2021neural,pang2021fewshot}, including~\cite{snarf,peng2021animatable,su2021anerf,2021narf} which are based on Neural Radiance Fields~\cite{mildenhall2020nerf}.

%% file: 3_method.tex
\begin{figure*}
    \vspace{-0.1cm}
    \includegraphics[width=\linewidth]{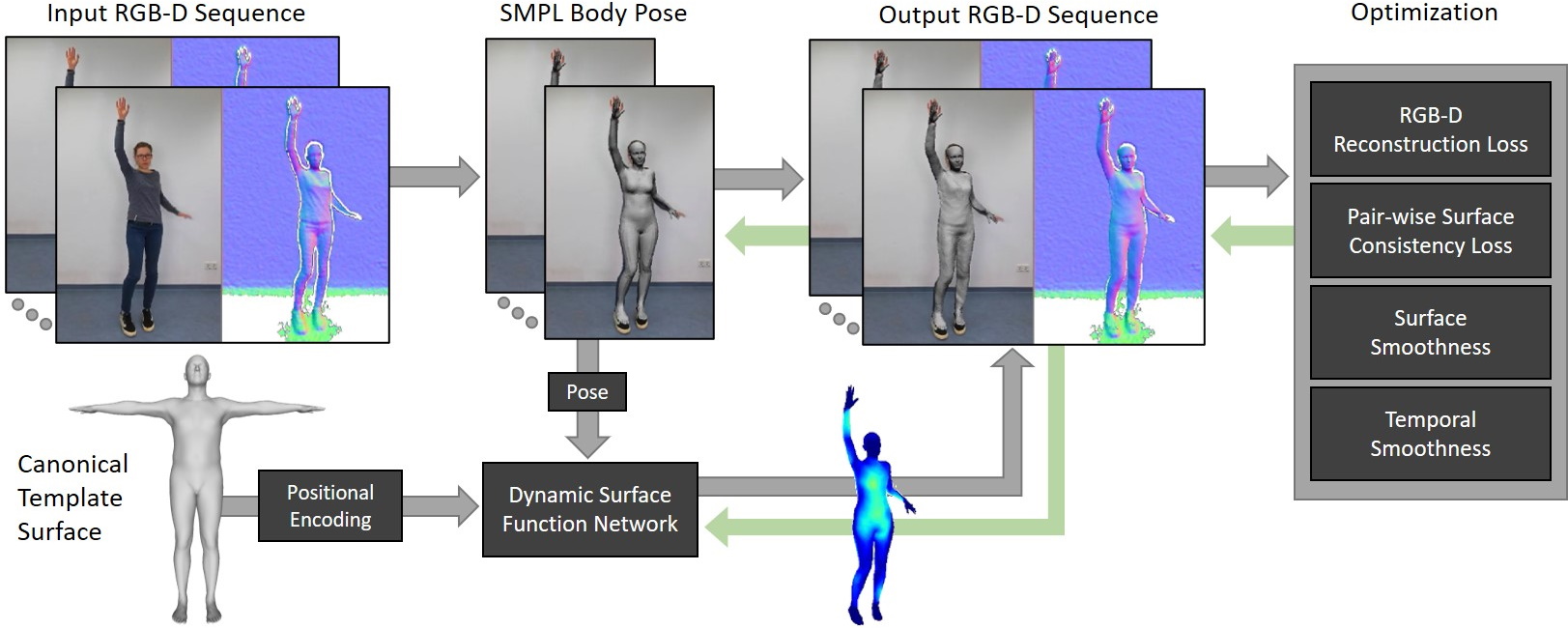}%
    \vspace{-0.3cm}
    \caption{
    Given an RGB-D input sequence, we jointly optimize for the SMPL body parameters and a pose-conditioned offset surface function (Dynamic Surface Function Network) which is represented by an MLP.
    Specifically, the MLP gets a positionally encoded surface point of the canonical template surface and the pose parameters as input and predicts the offset surface location.
    The pose parameters and the weights of the network are optimized based on the optimization of a global analysis-by-synthesis energy formulation (green arrow).
    }
    \label{fig:pipeline}
\end{figure*}

\section{Method}

We formulate the reconstruction of the human body as an energy optimization problem, globally, across the entire input sequence.
To model deforming clothing, we use an explicit offset surface function that is embedded on the SMPL surface.
This offset surface is represented as a multi-layer perceptron (MLP).
For an overview of our method refer to Fig.~\ref{fig:pipeline}.

\subsection{Model Definition}
Our method is built upon the SMPL body model~\cite{SMPL2015} and extends it to handle non-rigid offsets represented as a multi-layer perceptron (MLP).

\paragraph{Constrained SMPL model}
The SMPL model is parametrized by the shape parameters $\beta \in \mathbb{R}^{10}$ and the joint parameters $\delta \in \mathbb{R}^{72}$.
We use a hard constraint to enforce the parameters to stay in a predefined range.
That is, we use a differentiable mapping function $\delta(x)$ that maps our unconstrained joint parameters $x \in \mathbb{R}^{72}$, in an element-wise fashion, to a fixed range of the SMPL parameters:
\small
\begin{equation*}
    \delta(x) = \delta_{min} + \frac{\tanh(x)+1}{2} \cdot (\delta_{max} - \delta_{min}).
\end{equation*}
\normalsize
Note that we use axis-angle representation. Thus, we constrain the angle-scaled axis to be in a min-max bounding box.
We compute $\delta_{min}$ and $\delta_{max}$ from the MOSH dataset~\cite{mosh}.
In our experiments, this hard constraint was effective in stabilizing the tracking without the need of any additional regularization, e.g., using VPoser~\cite{SMPL-X:2019} (especially, for scenarios with more occlusions) or pose-conditioned joint angle limits~\cite{Akhter:CVPR:2015}.

\paragraph{Dynamic Offset Surface}
To model the dynamic offset surface on top of the SMPL model, we employ an MLP which is conditioned on the reparameterized SMPL parameters $x$.
Specifically, we embed the offsets $O_{\Theta}(v,x)$ in the canonical pose of the SMPL template surface $T$.
The offset of the surface point $v$, given the pose parameters $x$ is computed using a $8$-layer ReLU-MLP with $256$ feature channels per intermediate layer ($\Theta$ being the learnable parameters of the MLP).
We apply positional encoding~\cite{mildenhall2020nerf} to the input point $v$ before feeding it into the MLP (using $10$ frequencies).
The pose conditioning is concatenated to the point input in matrix form. Note that we do not use the pose of the root joint into consideration.
This representation of the pose conditioning is similar to the linear pose-dependent correctives of the SMPL model.
Note that the 3D space allows us to have a continuous input to the MLP without the need of handling texture seams, distortions or alike.
To rasterize the surface represented by the MLP, we sample the surface at surface points using a subdivided SMPL topology.

\paragraph{Model}
Given the constrained SMPL model and the dynamic offset model, we represent the vertices of a body as:
\small
\begin{equation*}
    V(\beta, x, \Theta) = LBS( V_{SMPL}(\beta, \delta(x)) + O_{\Theta}(T, x), \delta(x))
\end{equation*}
\normalsize
$LBS(V',\delta)$ is the linear blend skinning function that applies the rotations defined by $\delta$ to the 'unposed' vertices $V'=V_{SMPL}+O$.
$V_{SMPL}(\beta, \delta(x))$ computes the unposed SMPL model surface based on the shape PCA and the pose-dependent corrective space (see SMPL~\cite{SMPL2015}).
In our experiments, we exclude the hands and feet from the optimization (see appendix).

\subsection{Fitting Energy Formulation}

Given an RGB-D video sequence with $N+1$ frames, we minimize the following global energy:
\small
\begin{equation*}
    E_{Seq}(\mathcal{P}) = \sum_{i=0}^{N} E_{F}^i(\mathcal{P})
    + \sum_{i=1}^{N-1} E_{T}^i(\mathcal{P})
    + \sum_{i=0}^{N}\sum_{j=0}^{N} E_{C}^{i,j}(\mathcal{P})
    \label{eq:main}
\end{equation*}
\normalsize
This energy formulation considers per-frame energies $E_{F}^i$, temporal energies $E_{T}^i$, and pair-wise surface consistency constraints $E_{C}^{i,j}$.
$\mathcal{P}$ is the set of unknowns; namely the MLP weights $\Theta$, the shape $\beta$ and the per frame pose parameters $x_i$.

\paragraph{Per-frame Energy}

The per-frame energy is based on data-terms using the color frame $\mathcal{C}_i$ and depth frame.
Given the intrinsics of the depth camera, we back-project the depth maps into the camera space using the pinhole camera model $\Pi$, resulting in 3D locations per pixel which we call $\mathcal{D}_i$.
Based on the color frame $\mathcal{C}_i$, we estimate the 2D joint positions $\mathcal{J}_i^{OP}$ using OpenPose~\cite{openpose} and dense correspondences $\mathcal{M}_i^{DP}$ using DensePose~\cite{densepose}.
In addition, we use Graphonomy~\cite{Gong2019Graphonomy} to estimate the silhouette $\mathcal{S}^{G}_i$ of the person in the input image.
With these inputs, we define the per-frame energy as:
\small
\begin{align*}
    E_{F}^i(\mathcal{P}) &= w_{OP} \cdot E_{OpenPose}^i(\mathcal{P}) 
                                + w_{DP} \cdot E_{DensePose}^i(\mathcal{P}) \nonumber\\
                                &+ w_{Proj} \cdot E_{Projective}^i(\mathcal{P})
                                + w_{Sil} \cdot E_{Silhouette}^i(\mathcal{P}) \nonumber\\
                                &+ w_{Reg} \cdot E_{Reg}^i(\mathcal{P})
\end{align*}
\normalsize
The detected 2D joint locations are used as a sparse energy term:
\scriptsize
\begin{equation*}
    E_{OpenPose}^i(\mathcal{P}) = 1/K_J \cdot | \mathcal{J}_{i,j}^{OP} - \Pi(\mathcal{J}_{i,j})|
\end{equation*}
\normalsize
where $K_J=25$ is the number of joints and $\mathcal{J}_i$ are the corresponding regressed joints of the SMPL model.
The per-pixel DensePose energy term $E_{DensePose}^i$ is defined as:
\scriptsize
\begin{equation*}
    E_{DensePose}^i(\mathcal{P}) = \sum_{(p, c) \in \mathcal{M}_i^{DP}} \frac{P2P(V_c, \mathcal{D}_i(p))}{K_{DP}}
\end{equation*}
\normalsize
$K_{DP}$ is the number of valid correspondences $\mathcal{M}_i^{DP}$ estimated by DensePose ($p$ is the pixel and $c$ the correspondence on the SMPL surface).
$P2P(V_c, xyz)$ measures the $\ell_1$ point-to-point distance from the observation $xyz$ to the surface point of the model $V_c = Sample(c, V(\beta,x_i,\Theta))$.
The function $Sample(c, V(\beta,x,\Theta))$ computes the surface point based on the vertex indices ($i_0,i_1,i_2$) and barycentric coordinates ($b_0,b_1,b_2$) provided by the correspondence $c=(i_0,i_1,i_2,b_0,b_1,b_2)$.
In addition to the DensePose term, we use a dense data term using projective correspondences $\mathcal{M}_i^{Proj}$.
The projective energy term is defined as:
\scriptsize
\begin{align*}
    E_{Projective}^i(\mathcal{P}) = \sum_{(p, c) \in \mathcal{M}_i^{Proj}}  P2P(V_c, \mathcal{D}_i(p)) + N2N(V_c, \mathcal{D}_i(p)) \\
                                                                            + P2N(V_c,\mathcal{D}_i(p)) 
\end{align*}
\normalsize
$N2N(V_c, \mathcal{D}_i(p))$ measures the cosine similarity of the normals from the observation point $\mathcal{D}_i(p)$ to the surface point $V_c$.
$P2N(V_c, \mathcal{D}_i(p))$ measures point to plane distance between the observation point $\mathcal{D}_i(p)$ and the tangent plane at the source point $V_c$.

The per-frame silhouette energy term $E_{Silhouette}^i$ computes the difference between the mask predictions from ~\cite{Gong2019Graphonomy} and the rendered subject's silhouette $\mathcal{S}_i(\mathcal{P})$. It is defined as:
\scriptsize
\begin{equation*}
    E_{Silhouette}^i(\mathcal{P}) = | \mathcal{S}^{G}_i - \mathcal{S}_i(\mathcal{P}) |
\end{equation*}
\normalsize
Note that our differentiable rasterizer explicitly applies edge sampling at the boundary of the silhouette of the surface to compute gradients.

To optimize for a smooth surface, we use the regularizer $E_{Reg}^i(\mathcal{P})$.
Based on the topology of the template surface $T$, we apply a Laplacian regularizer:
\scriptsize
\begin{equation*}
    E_{Regularizer}^i(\mathcal{P}) = \sum_{m \in T} \left| V_m - \frac{\sum_{n \in neigh_m} V_n}{|neigh_m|} \right|^2 
\end{equation*}
\normalsize
Here, $neigh_m$ denotes the 1-ring neighborhood of the $m$-th vertex of the template topology.

\paragraph{Temporal Energy}

The temporal regularizer $E_{T}^i(\mathcal{P})$ is defined as:
\scriptsize
\begin{equation*}
    E_{T}^i(\mathcal{P}) = w_T^{surf} \cdot E_{T_{surf}}^i(\mathcal{P}) + w_T^{rot} \cdot E_{T_{rot}}^i(\mathcal{P}) 
\end{equation*}
\normalsize
We use a Laplacian regularizer on the surface of the model in the temporal domain:
\scriptsize
\begin{equation*}
    E_{T_{surf}}^i(\mathcal{P}) = | V_i - (V_{i-1} + V_{i+1}) / 2 |^2
\end{equation*}
\normalsize
In addition, we apply a temporal regularizer on the joint rotation matrices $R_i=R(\delta(x_i))$:
\scriptsize
\begin{equation*}
    E_{T_{rot}}^i(\mathcal{P}) = | R_i - R_{i-1} |^2 + | R_i - R_{i+1} |^2
\end{equation*}
\normalsize

\paragraph{Pair-wise Surface Consistency}

In each optimization step for each frame $i$, we randomly select another frame $j$ from the sequence to measure surface consistencies.
Specifically, we measure the difference of the offset surfaces:
\begin{equation*}
    E_{C}^{i,j}(\mathcal{P}) = w_C \cdot \omega(x_i,x_j) | \mathcal{V}_j \cdot ( O_{\Theta}(T, x_i)) - O_{\Theta}(T, x_j)) |
\end{equation*}
$\omega(x_i,x_j)=exp(-|R(\delta(x_i))-R(\delta(x_j))|^2)$ measures the similarity of the poses in the two frames (excluding the root joint pose) and $\mathcal{V}_j$ denotes the surface visibility in frame $j$.

\subsection{Optimization Scheme}

The energy formulation in Eq.~\ref{eq:main} is optimized in two stages.
Specifically, we first minimize the energy only considering the SMPL body parameters for shape and pose to get a good initial estimate.
We use L-BFGS~\cite{lbfgs} to optimize for these shape and pose parameters in sequential order (we initialize the parameters with the parameters of the previous frame).
Using this initial fitting, we optimize for all parameters $\mathcal{P}$ in a joint optimization using ADAM~\cite{adam} with random sampling of the input images.
We refer to the appendix for the used hyperparameters.

%% file: 4_results.tex
\section{Results}

\begin{figure}
    \centering
    \includegraphics[width=0.97\linewidth]{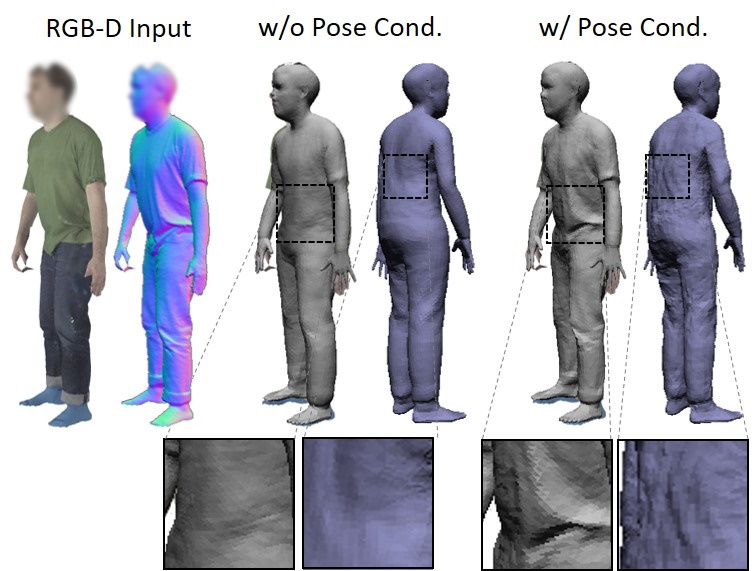}
    \caption{
        Without pose conditioning, only a static surface function is learned, and dynamically changing pose-dependent details such as wrinkles are not reconstructed. In blue, we show the view rotated by $180^{\circ}$.
    }
    \label{fig:static_vs_dynamic}
\end{figure}

\begin{figure*}
    \centering
    \includegraphics[width=\linewidth]{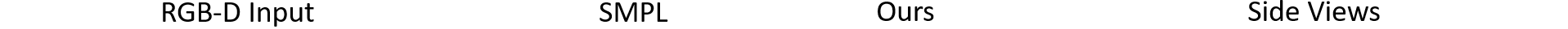}
    \vspace{0.15cm}
    \includegraphics[width=\linewidth]{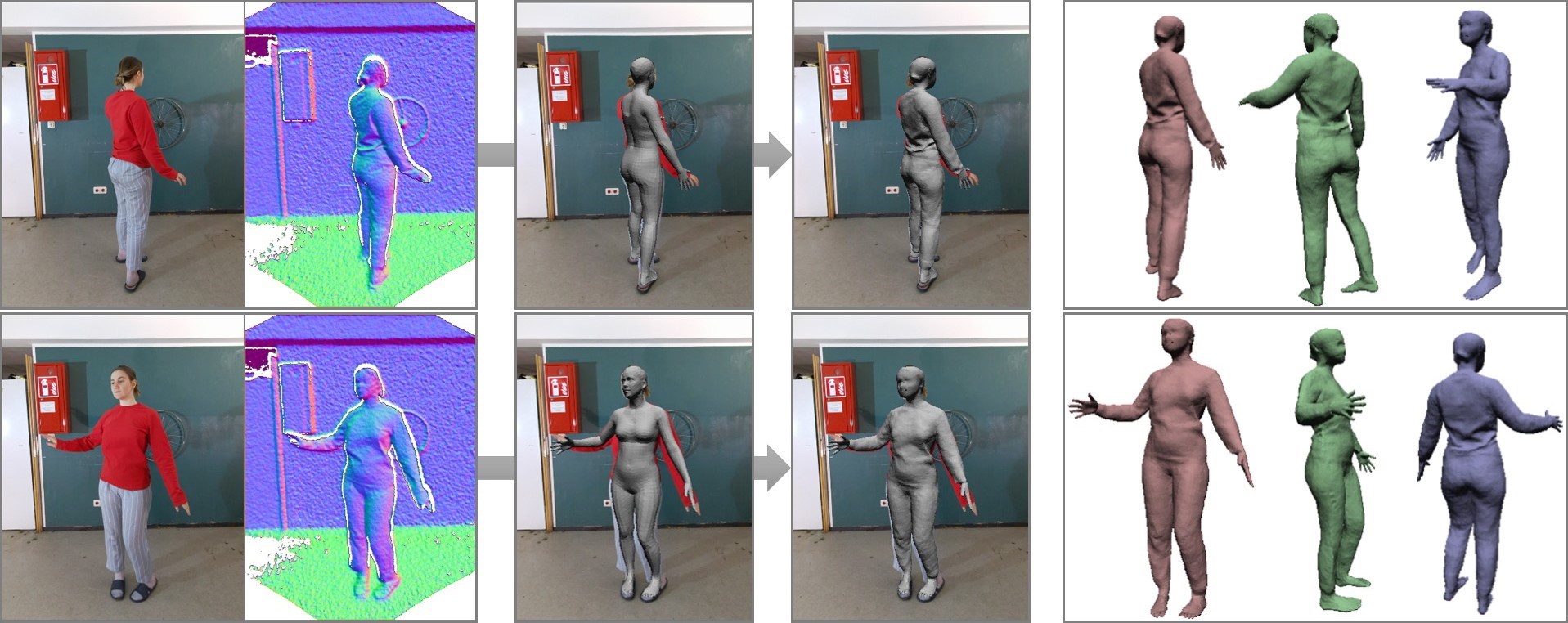}
    \vspace{0.15cm}
    \includegraphics[width=\linewidth]{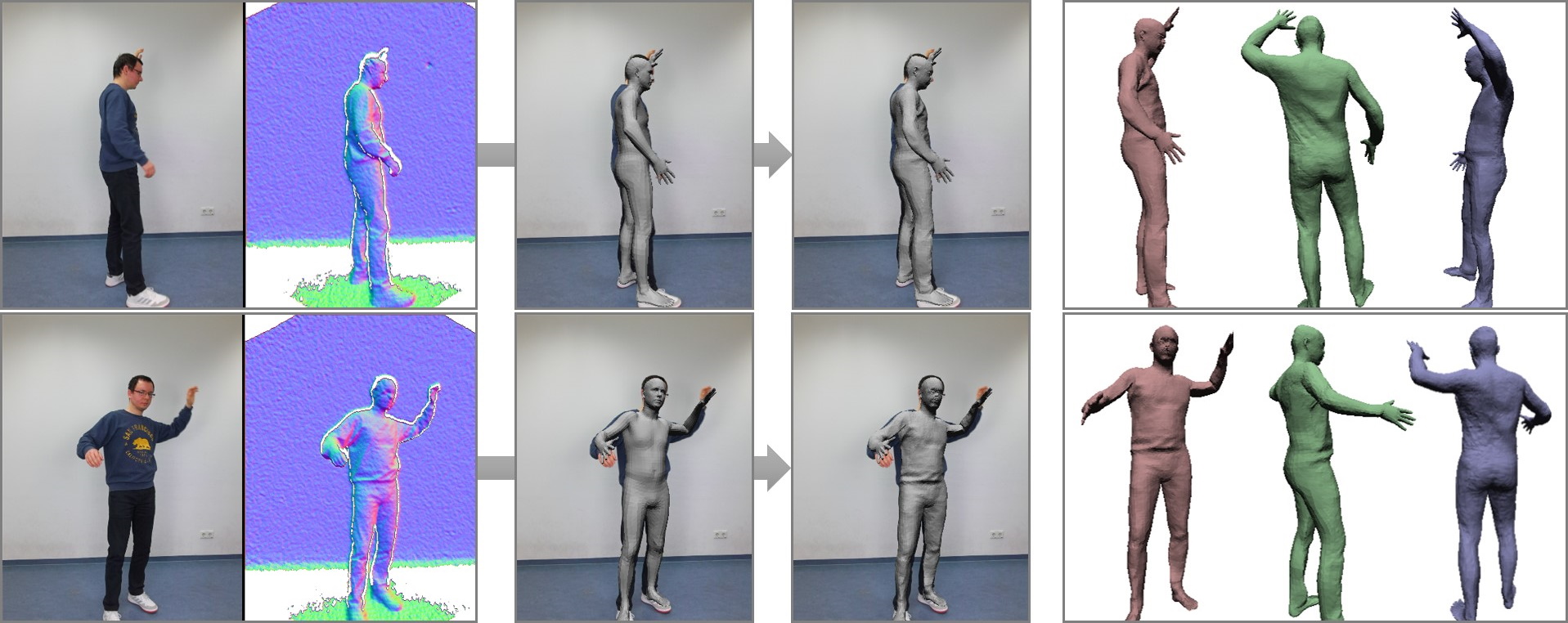}
    \includegraphics[width=\linewidth]{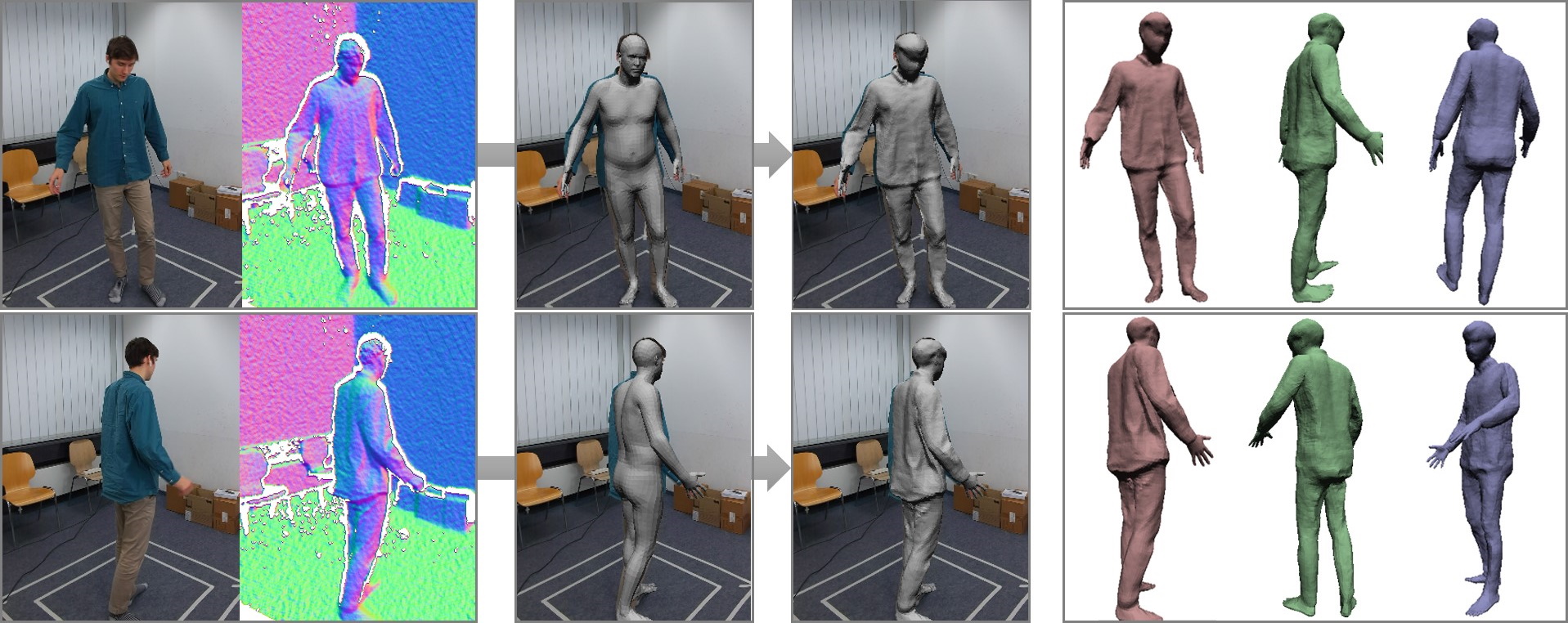}
    \caption{
        Temporally coherent reconstructions of clothed humans using our proposed dynamic surface function networks.
        The input data has been captured by a Microsoft Kinect Azure.
        Each sequence is $200$ frames long.
    }
    \label{fig:results}
\end{figure*}

In this section, we evaluate our method on synthetic as well as on real-world data.
For quantitative evaluations, we use synthetic sequences based on the BUFF~\cite{buff} dataset.
Real data inputs are captured with a Microsoft Kinect Azure at a depth resolution of $640$x$576$ and color image resolution of $1080p$ at $15$ fps (each recording is $200$ frames long).%
We align color images with depth maps using OpenCV and apply distortion correction.
These sequences are used to demonstrate the applicability of our method to real-world data, as well as to show qualitative comparisons.
Specifically in Fig.~\ref{fig:results}, we show reconstructions of different persons in varying clothing.
As can be seen, we faithfully reproduce the input data including the pose-dependent deformations of the surface.
The temporal coherence of our reconstructions can be seen in the supplemental video.
In Fig.~\ref{fig:teaser}, we show results for novel poses using our representation based on poses from the MOSH dataset~\cite{mosh}.

\subsection{Ablation Studies}

The key components of our approach are the temporally consistent tracking of the SMPL body as well as the reconstruction of the surface using a pose-dependent MLP.
In the following, we will analyze the effects of our optimization scheme as well as the dynamic surface MLP.

\vspace{-0.2cm}
\paragraph{Static vs. Dynamic Surface MLP}
To handle pose dependent deformation of the geometry such as wrinkles of the clothing, we use a dynamic surface function network.
This network is conditioned on the pose parameters of the underlying SMPL model.
In Fig.~\ref{fig:static_vs_dynamic}, we show a comparison of using a dynamic surface function network and a static one (i.e., without providing the pose conditioning to the network).
As can be seen, only the pose conditioned network is able to represent the dynamically deforming surface.

\vspace{-0.2cm}
\paragraph{Pair-wise Surface Consistency}

\begin{figure}
    \vspace{-0.1cm}
    \includegraphics[width=\linewidth]{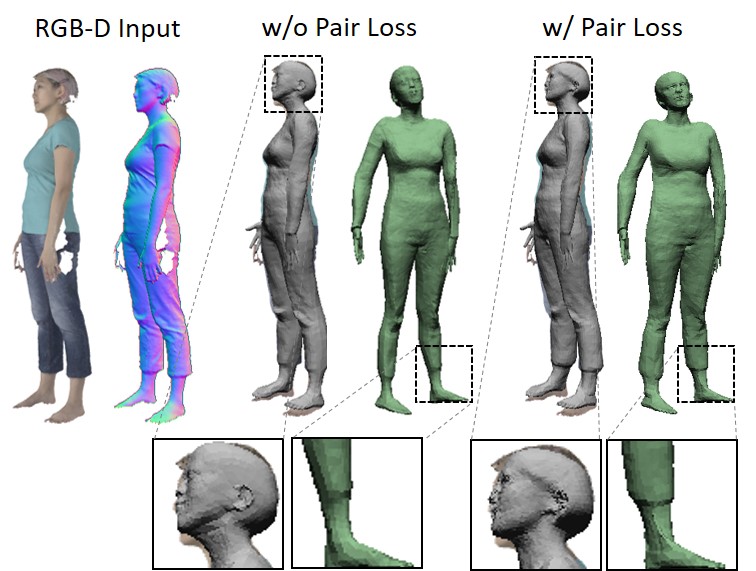}
    \caption{
        To reconstruct a consistent surface, we apply a pair-wise surface consistency loss, which measures the surface difference of pairs of frames weighted by their pose similarity.
        In green, we show the view rotated by $90^{\circ}$.
    }
    \label{fig:pairs}
\end{figure}

The pair-wise consistency regularizes the surface network to predict similar surfaces if the pose is similar.
In Fig.~\ref{fig:pairs}, we show the effect of this regularizer.
As can be seen, the pair-wise surface consistency loss leads to more details as well as to a 3D consistent surface reconstruction.
Without the pair-wise loss, the surface tends to be closer to the SMPL surface.

\vspace{-0.2cm}
\paragraph{Joint Optimization}

Our method is based on a joint global optimization, i.e., the MLP is trained in conjunction to the SMPL model parameters.
In Tab.~\ref{tab:buff}, we list the quantitative evaluation for an ablation study that uses sequential optimization (first optimization of the SMPL parameters, and then training of the MLP).
As can be seen, joint optimization is crucial for the reconstruction quality.

\subsection{Comparisons}

\begin{figure}
    \centering
    \vspace{-0.5cm}
    \includegraphics[width=\linewidth]{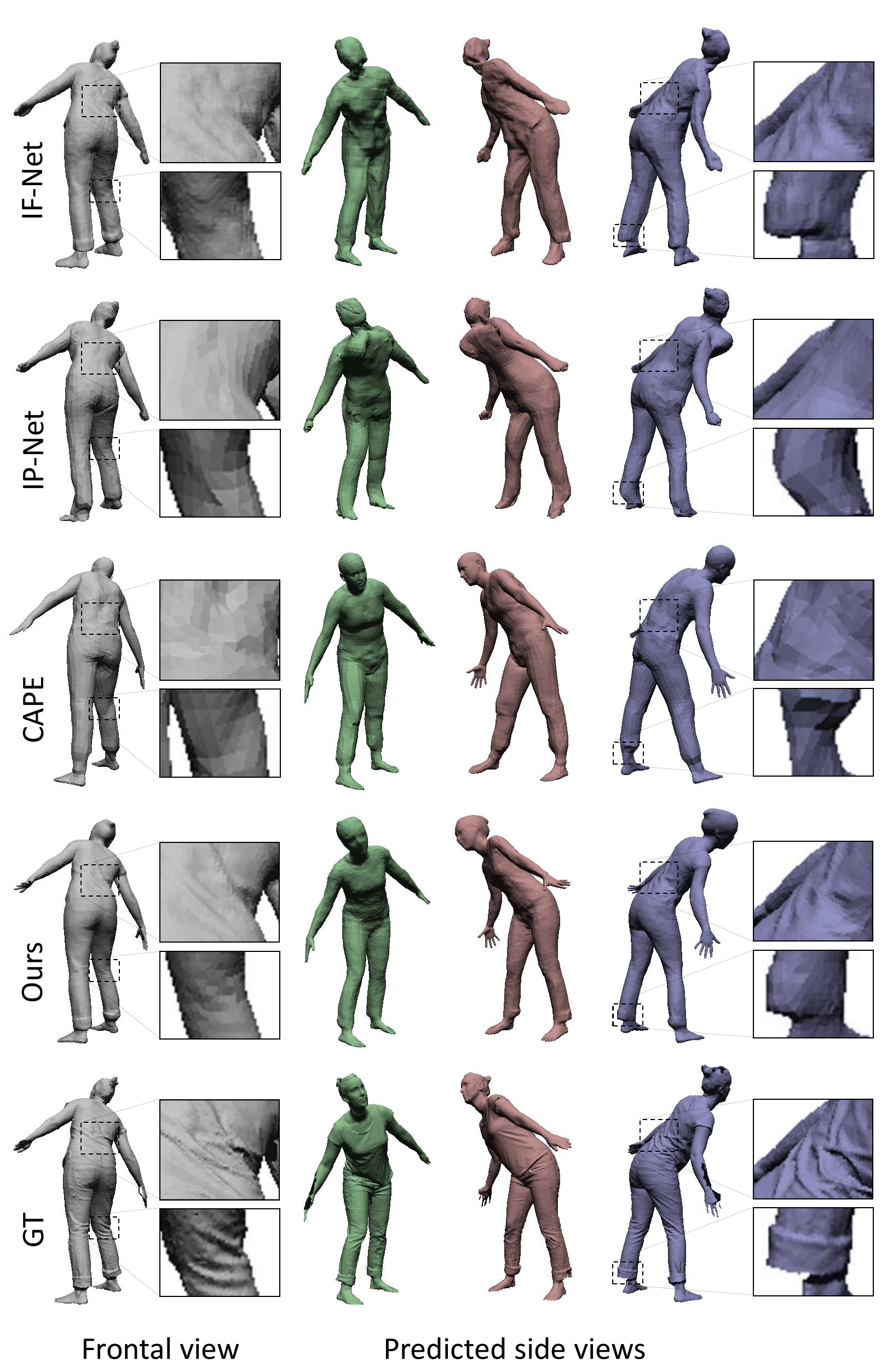}
    \vspace{-0.65cm}
    \caption{
        Our model is an explicit surface function. CAPE~\cite{CAPE} also models an explicit surface, it predicts the offsets from SMPL with a graph based decoder. In contrast, IF-Nets~\cite{chibane20ifnet} represent the surface implicitly using an occupancy function. IP-Net~\cite{bhatnagar2020ipnet} is a hybrid approach that fits the SMPL model with additional vertex displacements to an estimated implicit function.
    }
    \vspace{-0.5cm}
    \label{fig:comparison_implicit_function}
\end{figure}

\paragraph{Implicit Surface Representations}

Our dynamic surface function networks estimate the surface points explicitly. A classical rasterizer can be used to render these surfaces.
In contrast, implicit surface representations (like an occupancy function) need to be ray-casted or converted into an explicit surface representation~\cite{marching_cubes}.
Since implicit functions like IF-Nets~\cite{chibane20ifnet} do not provide explicit correspondences over a time series, regularizing the surface to deform in a smooth and temporally consistent manner is non-trivial (regions can disappear and appear at a different location).
In Fig.~\ref{fig:comparison_implicit_function}, we show a comparison to IF-Nets and IP-Nets on an RGB-D sequence.
IP-Net~\cite{bhatnagar2020ipnet} is a hybrid, leveraging the reconstruction abilities of IF-Nets and the controllability of the SMPL body model.
Both IF-Nets and IP-Nets take a point cloud of a single frame as input and predict the implicit surface.
We use the masks predicted by Graphonomy to remove the background points from the input.
This input is fed into the pretrained networks for single views provided by the authors as shown in their publications.
Note that the authors explicitly state that the networks generalize to continuous articulations of temporal data from new data sources.
In Tab.~\ref{tab:buff}, we show the corresponding errors w.r.t. IoU, chamfer-distance and normal consistency.
We observe that our method leads to better reconstructions, with a temporally consistent output mesh (see supplemental video).

\begin{table}[]
    \centering
    \begin{tabular}{|l|c|c|c|c|}
        \hline
        \textbf{Method} & \textbf{IoU} $\uparrow$ & \textbf{C-$\ell_2$} $\downarrow$ & \textbf{NC} $\uparrow$ \\ %
        \hline

        IF-Nets~\cite{chibane20ifnet} & 0.818 &	1.8cm &	0.903 \\ \hline
        IP-Nets~\cite{bhatnagar2020ipnet} & 0.783 &	2.1cm &	0.861 \\ \hline
        CAPE~\cite{CAPE} & 0.648 &	2.5cm &	0.844 \\ \hline \hline
        Ours w/o joint optimization & 0.687 &	2.4cm &	0.877 \\ \hline
        Ours & \textbf{0.832} &	\textbf{1.6cm} &	\textbf{0.916} \\ \hline

    \end{tabular}
    \vspace{-0.125cm}
    \caption{Quantitative comparisons based on a sequence of the BUFF dataset~\cite{buff} (see Fig.~\ref{fig:comparison_implicit_function}). For all methods, we provide synthetically rendered monocular RGB-D data inputs. In the table, we report the numbers for IoU, chamfer distance (using $\ell_2$-norm) and normal consistency w.r.t. the complete meshes from the dataset. }
    \label{tab:buff}
\end{table}

\begin{table}[]
    \centering
    \begin{tabular}{|l|c|}
        \hline
        \textbf{Method} & \textbf{EPE} $\downarrow$ \\
        \hline
        BodyFusion~\cite{BodyFusion} & 2.77cm \\ \hline
        IP-Nets~\cite{bhatnagar2020ipnet} & 14.17cm$^*$ \\ \hline
        CAPE~\cite{CAPE} & 5.51cm \\ \hline
        SMPL & 4.33cm \\ \hline \hline
        Ours & \textbf{2.63cm} \\ \hline
    \end{tabular}
    \vspace{-0.125cm}
    \caption{
            Quantitative comparisons based on the BodyFusion dataset~\cite{BodyFusion} which provides tracked VICON marker locations for each RGB-D frame.
            The mean end-point-error (EPE) is measured in 3D based on an $\ell_1$ distance between the tracked VICON markers and the corresponding points on the reconstructed mesh. Note that both the CAPE baseline and our method use the SMPL tracking as initialization. $*$ not using temporal information.
            }
            \vspace{-0.25cm}
    \label{tab:bodyfusion}
\end{table}

\begin{figure}
    \includegraphics[width=\linewidth]{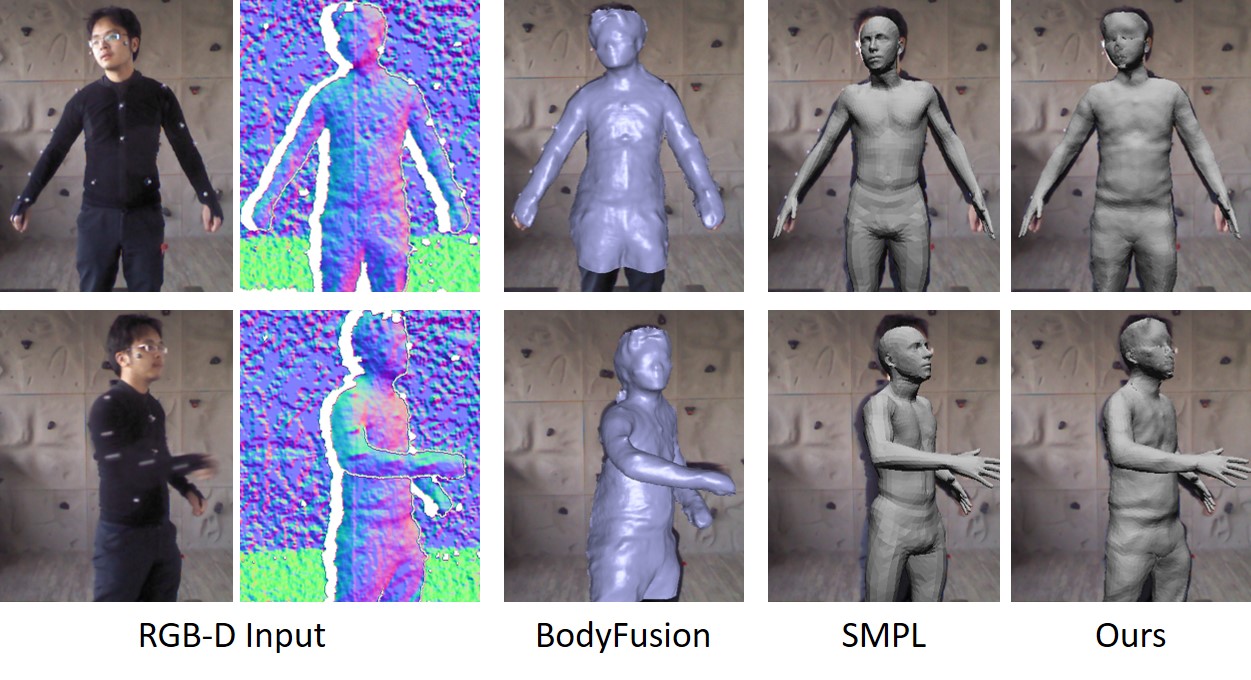}
    \vspace{-0.75cm}
    \caption{
        Qualitative comparison to BodyFusion~\cite{BodyFusion} based on a sequence of their dataset.
        Note that this dataset has low quality inputs, but still our approach is able to integrate details over the sequence to reconstruct a high quality model.
    }
    \label{fig:body_fusion}
\end{figure}

BodyFusion~\cite{BodyFusion} is a fusion method that integrates the depth measurements into a volumetric SDF representation.
It leverages the skeleton of a body as a kinematic prior, thus, outperforming general non-rigid fusion methods like DynamicFusion~\cite{Newcombe_2015_CVPR} or VolumeDeform~\cite{innmann2016volume} in the scenario of body reconstruction.
In Tab.~\ref{tab:bodyfusion}, we quantitatively compare our method to BodyFusion as well as to IP-Nets.
Note that both BodyFusion as well as our method are using temporal information, while IP-Net is applied frame-by-frame, thus, leading to the highest tracking error.
The tracking error is measured by a mean $\ell_1$ end-point error on the dataset provided by the authors of BodyFusion~\cite{BodyFusion}.
Aside from the low-quality depth and color images, the dataset provides the coordinates of tracked VICON markers used for evaluation (see Fig.~\ref{fig:body_fusion}).
Our method results in the lowest error and qualitatively results in sharper results.

\vspace{-0.2cm}
\paragraph{Explicit Surface Representations}

In the results discussed above, we already mention IP-Nets~\cite{bhatnagar2020ipnet} which model the surface explicitly with per-vertex displacements on top of the SMPL model, also known as SMPL-D.
Our approach with an MLP \textit{without} pose conditioning is closely related to SMPL-D (see Fig.~\ref{fig:static_vs_dynamic}). Instead of a discrete number of displacement vectors, we represent the surface using a continuous function that can be evaluated on any surface point of the SMPL model.
In Tab.~\ref{tab:buff}, we also include a comparison to the graph neural network-based GAN for clothing CAPE~\cite{CAPE}.
We used our optimization framework to estimate the latent codes, based on the RGB-D inputs (see appendix for more details).
While the generative approach has several advantages for reconstructing humans from partial views or RGB images, it does not capture the detail that our reconstruction method recovers.

\begin{figure}
    \centering
    \vspace{-0.25cm}
    \includegraphics[width=0.9\linewidth]{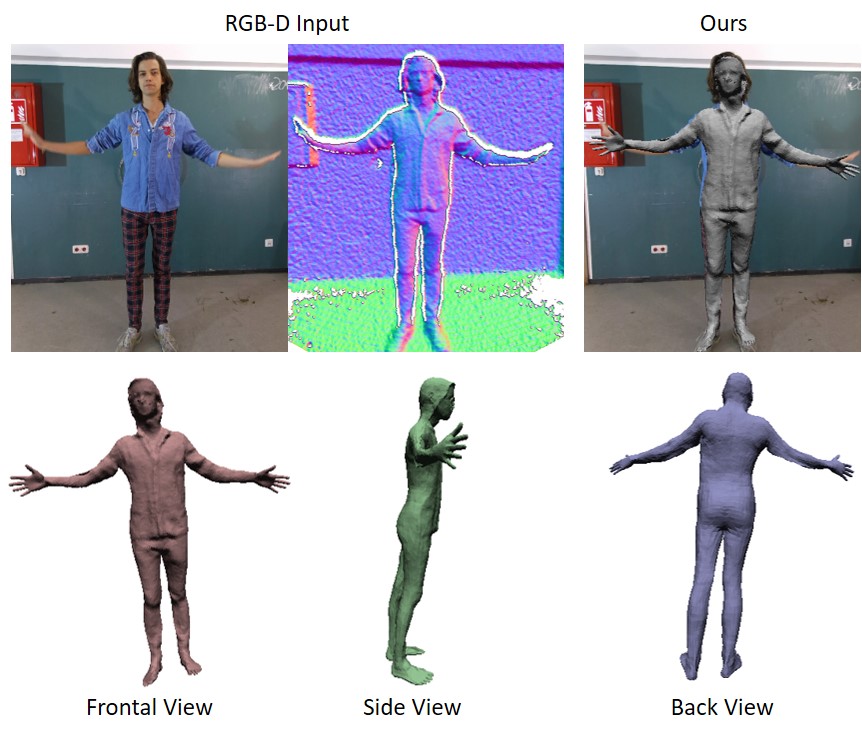}
    \vspace{-0.25cm}
    \caption{
        Our approach jointly optimizes the pose and the dynamic surface function network based on the given input data.
        It does not reconstruct regions that were not visible in the sequence.
        In the shown example, the input data only contained frontal views of the person.
    }
    \label{fig:limitations}
\end{figure}

%% file: 5_conclusion.tex
\section{Discussion}
Our Dynamic Surface Function Networks are  able to reconstruct and track a variety of sequences of different people.
The explicit representation of the surface allows us to use differentiable rasterization for rendering, and to generate a temporally consistent mesh as well as global correspondences, but it comes with the limitation of a fixed topology, and, thus, is not able to represent topologically changing surfaces.

Our approach is an optimization approach and belongs to the class of methods that must be trained for each new input sequence~\cite{mildenhall2020nerf,thies2019neural,sitzmann2019siren}.
In particular, for the reconstruction of a controllable representation this retraining of the dynamic surface function network is practical and crucial to get fine scale details such as wrinkles.
If regions of the body are not visible in the input sequence (i.e., no data terms), the surface in these regions is only regularized to be smooth (see Fig.~\ref{fig:limitations}).
The assumption of the pair-wise consistency loss (similar clothing geometry for similar poses) might not always hold for loose garment.
Extending our approach to such apparel is an interesting direction for future work.
Further improvements on wrinkle expressiveness might be achieved by leveraging the dense color input.

\section{Conclusion}

Our approach reconstructs and tracks a clothed human using a single commodity RGB-D sensor, obtaining a temporally-consistent surface reconstruction.
The underlying dynamic surface function network is able to represent pose-dependent deformations of the surface and allows to re-animate the body.
In our experiments, we demonstrate the effectiveness of this representation and show state-of-the-art reconstruction performance.
The proposed representation offers a variety of advantages such as a consistent mesh structure, dynamically changing surface and the possibility to extend it to other attributes (see appendix). %

%% file: 7_acknowledgments.tex
\section*{Acknowledgements}

The work is funded by Huawei, a  TUM-IAS  Rudolf  M\"o\ss{}bauer  Fellow-ship, the ERC Starting Grant Scan2CAD (804724), the German  Research  Foundation  (DFG)  Grant \textit{Making  Machine Learning on Static and Dynamic 3D Data Practical}, and the BMBF-funded Munich Center for Machine Learning.

%% file: 6_supplement.tex
\appendix

\section{Implementation Details}
\label{sec:impl}

\subsection{Network Architecture}

The dynamic surface function is represented as a multi-layer perceptron (MLP).
In our experiments, we use an 8-layer MLP with ReLU activation functions for the intermediate layer (each intermediate layer has a feature dimension of $256$).
The final output layer uses a tanh activation function, allowing us to specify a maximal amplitude of the offset surface (in our experiments 25cm).
The network architecture is inspired by Mildenhall et al.~\cite{mildenhall2020nerf}, using the positional encoding for the sample point coordinate input.
To represent pose-dependent deformations, we condition the dynamic surface function network also on pose parameters.
Specifically, we compute the 'pose feature' $\mathcal{F} = [\mathbf{F}_1,\ldots,\mathbf{F}_{23}] \in \mathbb{R}^{23 \times 9}$, where $\mathbf{F}_k = (R_k - \mathit{Id})$ is the feature component of a body part $k$ (the root part is not included).
This pose feature is describing the global pose of a human.
Since most deformations are local (e.g., the pose of the leg does not influence the surface of an arm), we compute a local pose conditioning of a sample point based on the linear blend-skinning weights of SMPL.
Specifically, we enable the pose conditioning of the corresponding joints defined by the SMPL skinning weights, as well as for the adjacent nodes (2-ring neighborhood, i.e., parent and grandparent node, as well as child and grandchild node): 
\small
\begin{equation*}
    \hat{\mathcal{F}} = (lbs_k \cdot \mathbf{N}_k) \cdot \mathcal{F},
\end{equation*}
\normalsize
where $lbs \in \mathbb{R}^{23}$ are the skinning weights of a sample point, $\mathbf{N} \in \mathbb{R}^{23 \times 23}$ the 2-ring adjacency matrix.

Note, during training we augment the pose conditioning $\mathcal{F}$ with noise to control overfitting.
Specifically, we apply additive normal distributed noise with a standard deviation of $0.1$.

\subsection{Optimization}

\paragraph{Optimizer settings}
The energy function is optimized in two stages. First, we fit the SMPL template to match the observations sequentially.
We apply the L-BFGS~\cite{lbfgs} optimizer with line search, history size of $20$ and $20$ maximum iterations.
We observe that using Adam~\cite{adam} at this stage is not efficient, since it struggles to reconstruct rotations in the axis-angle form.
The optimization is executed for $15$ passes through the dataset with a fixed learning rate of $0.1$ and then for another $15$ passes with the learning rate linearly decreasing to $0$.

During the second stage our objective is to reconstruct all parameters $\mathcal{P}$ jointly (MLP and SMPL parameters).
At this stage, we use a standard Adam optimizer with $(0.9, 0.999)$ blending weights for the first and second momentum respectively. 
The optimization is carried out on random samples from the sequence with the first $100$ global passes updated by a static learning rate of $0.00005$ and remaining $300$ passes by a linearly decaying learning rate.
As soon as the learning rate is starting to decrease, we enable the dynamic conditioning, to capture the pose specific clothing deformations from reconstructed subjects.

The optimization using ADAM takes approximately $60s$ per epoch ($200$ frames) while the initial fitting with L-BFGS takes around $700s$ per epoch.

\begin{figure}
    \centering
    \includegraphics[width=0.25\linewidth]{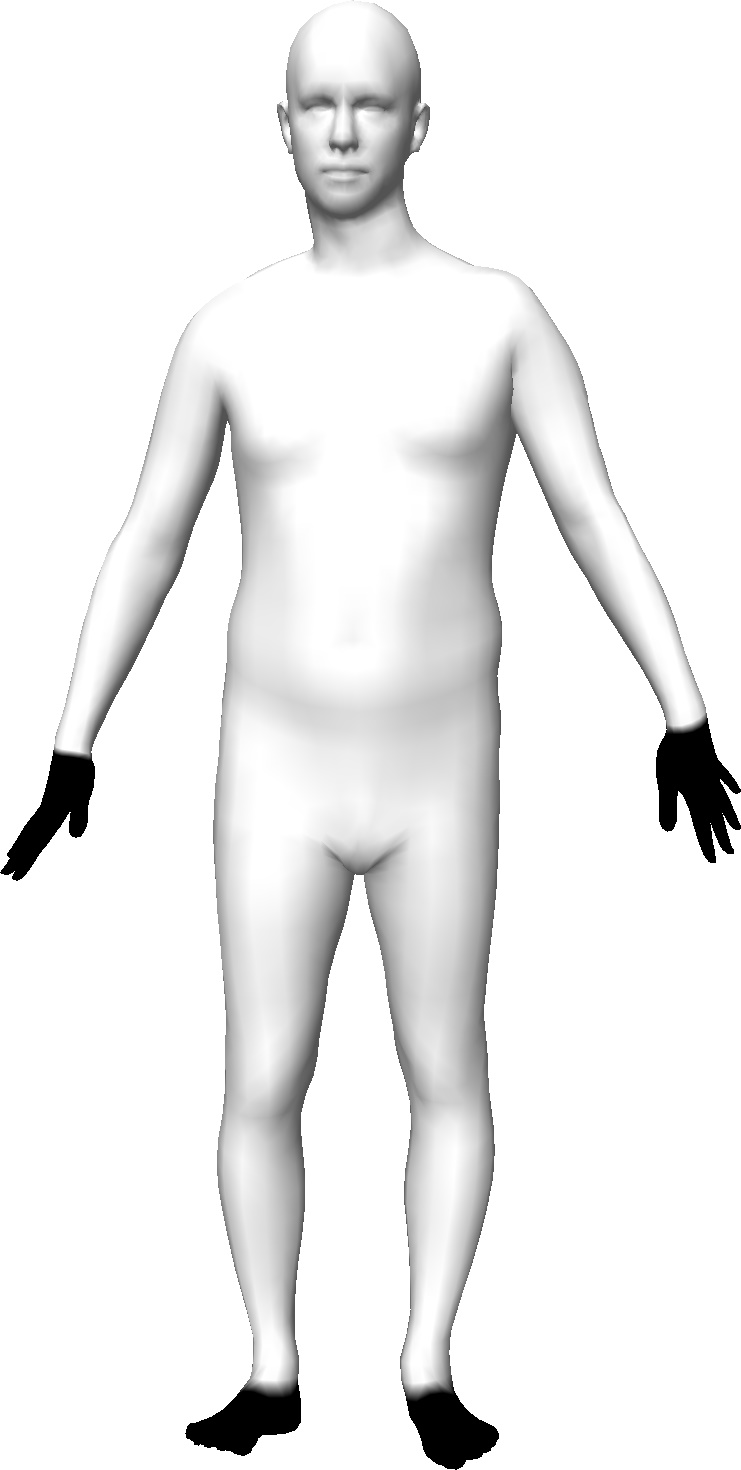}
    \caption{
        We exclude the hands and feet from optimization.
    }
    \label{fig:comparison_implicit_function}
\end{figure}

\paragraph{Loss weights}

\begin{table}[]
    \centering
    \resizebox{\linewidth}{!}{
    \begin{tabular}{|l|c|c|c|}
        \hline
        \textbf{Energy Term} & \textbf{Symbol} & \textbf{Value} & \textbf{Space}\\
        \hline
        Sparse OpenPose & $w_{OP}$ & 1500 & normalized image space \\ \hline
        Dense Densepose & $w_{DP}$ & 25 & 3D space in meters \\ \hline
        Dense Projective & $w_{Proj}$ & 100 & 3D space in meters \\ \hline
        Silhouette & $w_{Sil}$ & 50 & normalized image space \\ \hline
        Surface Smoothness & $w_{Reg}$ & 1500 & 3D space in meters \\ \hline
        \hline
        Temporal Smoothness & $w_{T}^{Surf}$ & 100 & 3D space in meters  \\ \hline
        Temporal Smoothness & $w_{T}^{Rot}$ & 15 & rotation matrices  \\ \hline
        \hline
        Pair-wise Consistency & $w_C$ & 15 & 3D space in meters \\ \hline
    \end{tabular}
    }
    \caption{
            Energy term weights during optimization.
            }
    \label{tab:weights}
\end{table}

As described in the main paper, our optimization is based on a set of different energy terms.
In Tab.~\ref{tab:weights}, we specify the used weights during optimization.
Note that we optimize in two stages as described above.
For the initial fitting of the SMPL parameters, we increase the OpenPose weight $w_{OP}$ to $10000$ and disable the projective energy term during the first two optimization iterations (since the body is not yet roughly aligned with the body in the image, thus, leading to wrong projective correspondences).
The temporal regularizers in this initial fitting procedure are turned on after the $5th$ pass.
Note that all terms are normalized by their respective number of residuals (i.e., by the number of pixels).
We prune projective correspondences based on distance ($0.5m$) and deviation in normals ($45^{\circ}$).

\subsection{Surface Sampling}

For rendering, we need to sample the surface.
We use the SMPL triangulation and subdivide it with a 1-to-4 subdivision scheme (each triangle is subdivided into 4).
Based on these samples and the corresponding topology, we evaluate the dynamic surface function network to retrieve the actual surface position.
These positions are then sent to the GPU rasterizer to render the surface, used for the analysis-by-synthesis process.
Note that correspondences from DensePose~\cite{densepose} lead to additional samples on the surface.

\subsection{Baseline Implementation}

In the main paper, we discuss results based on the CAPE cloth model~\cite{CAPE}.
We leverage our fitting pipeline to optimize the energy with respect to the latent codes of the CAPE model.
Specifically, we take the publicly available checkpoints for the \textit{male} and \textit{female} subjects (with clothing latent space of size $64$, pose condition size of $32$ and clothing type condition size of $32$) and define the objective as latent codes' optimization for the CAPE decoder.
In particular, we append the losses from the first stage of our optimization procedure to the Tensorflow~\cite{tensorflow} graph of the CAPE decoder, and initialize the reconstruction process with the parameters from the SMPL only optimization.

\section{Additional Results}
\label{sec:add_exp}

\subsection{Comparisons}

\begin{figure*}[t]
    \centering
    \includegraphics[width=0.9\linewidth]{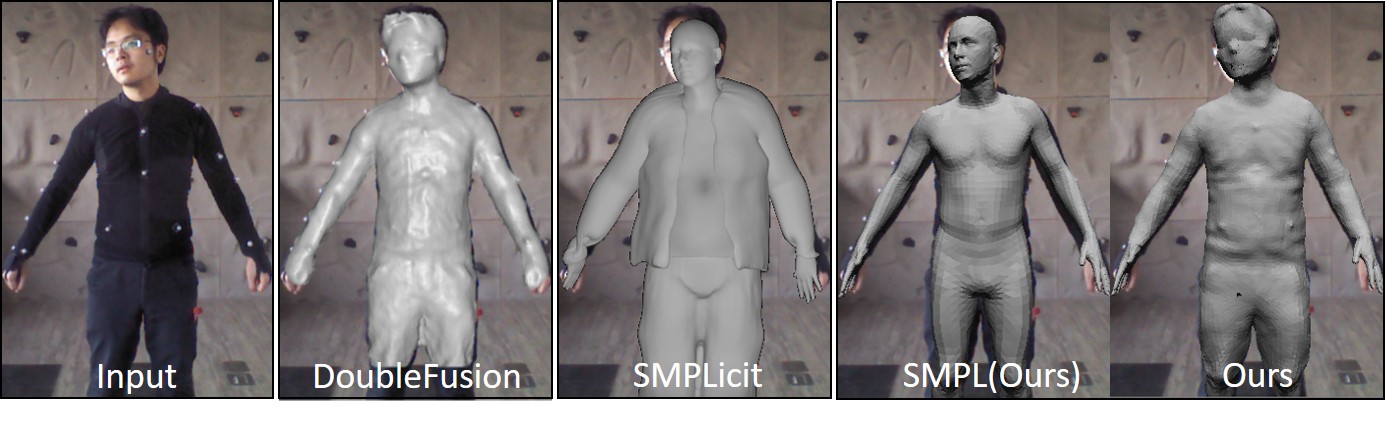}
    \vspace{-0.3cm}
    \caption{
        Additional qualitative comparison to DoubleFusion~\cite{DoubleFusionPAMI_2019} and SMPLicit~\cite{corona2021smplicit}.
        DoubleFusion is incrementally fusing the depth-observations to reconstruct the final body shape, while SMPLicit uses a generative approach to produce an output garment configuration that is close to the input.
        In contrast, our method globally optimizes for the actual shape.
    }
    \label{fig:doublefusion}
\end{figure*}

We provide an additional comparison to the incremental reconstruction method DoubleFusion \cite{DoubleFusionPAMI_2019} and to the topology-aware generative clothed human model SMPLicit~\cite{corona2021smplicit}. 
As can be seen in Fig.~\ref{fig:doublefusion}, our approach is able to reconstruct more details than DoubleFusion, especially in the face region and also on the body.
In contrast to DoubleFusion, our approach optimizes for a globally consistent representation that does not overfit to latest observations. 
The result produced by SMPLicit depicts a possible garment configuration for the specimen, however, as a method based on a generative model it does not match observed data as closely as an actual reconstruction method.

\subsection{Reconstructing Surface Colors}

\begin{figure}
    \centering
    \includegraphics[width=1.0\linewidth]{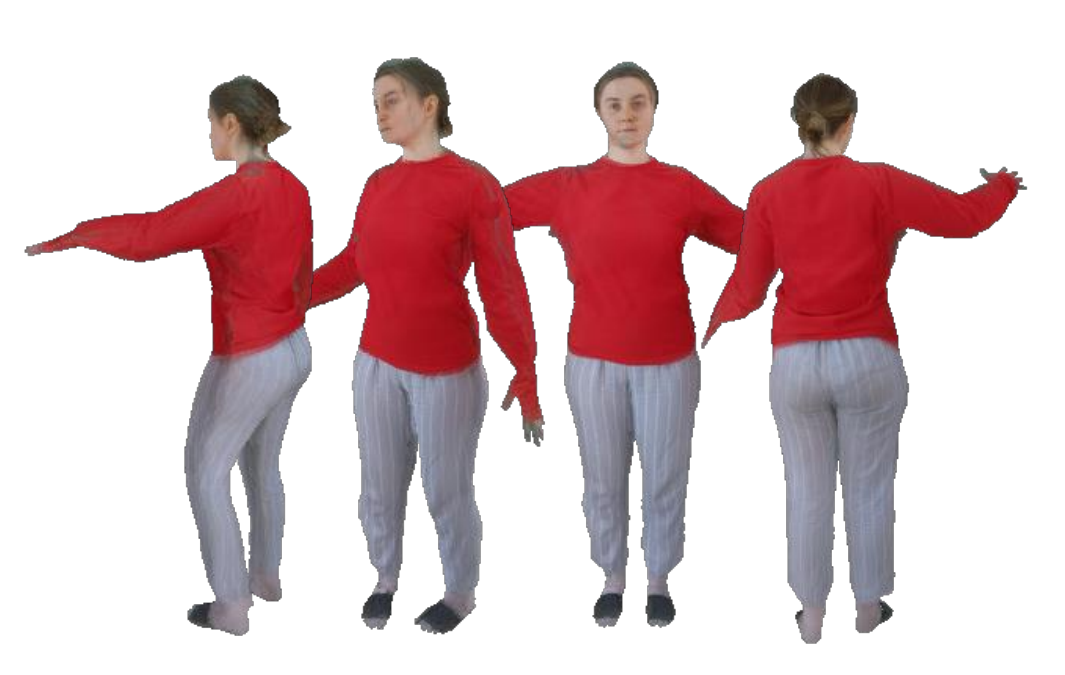}
    \caption{
        Our model can be extended to reconstruct the surface color. We use an MLP similar to the shape MLP to predict the color.
    }
    \label{fig:color_segmentations}
\end{figure}

In Fig.~\ref{fig:color_segmentations}, we show the result of optimizing an additional MLP for the surface color.
Specifically, we use the same architecture as the shape MLP and predict color values for each surface point.
The MLP used in this experiment has $6$ layers and a latent size of $256$.
We use $16$ frequency bands for the positional encoding.
We use a pretrained shape MLP, and train the color MLP with an $\ell_1$ reconstruction and perceptual losses \cite{Johnson2016Perceptual}.
This experiment shows, that you can easily reconstruct the surface appearance of the person within our framework.
Note that the incorporation of the color to refine the tracking and shape prediction is still open for follow-up works (i.e., joint optimization of the color and the shape MLP).

\begin{figure*}[t]
    \centering
    \includegraphics[width=1.0\linewidth]{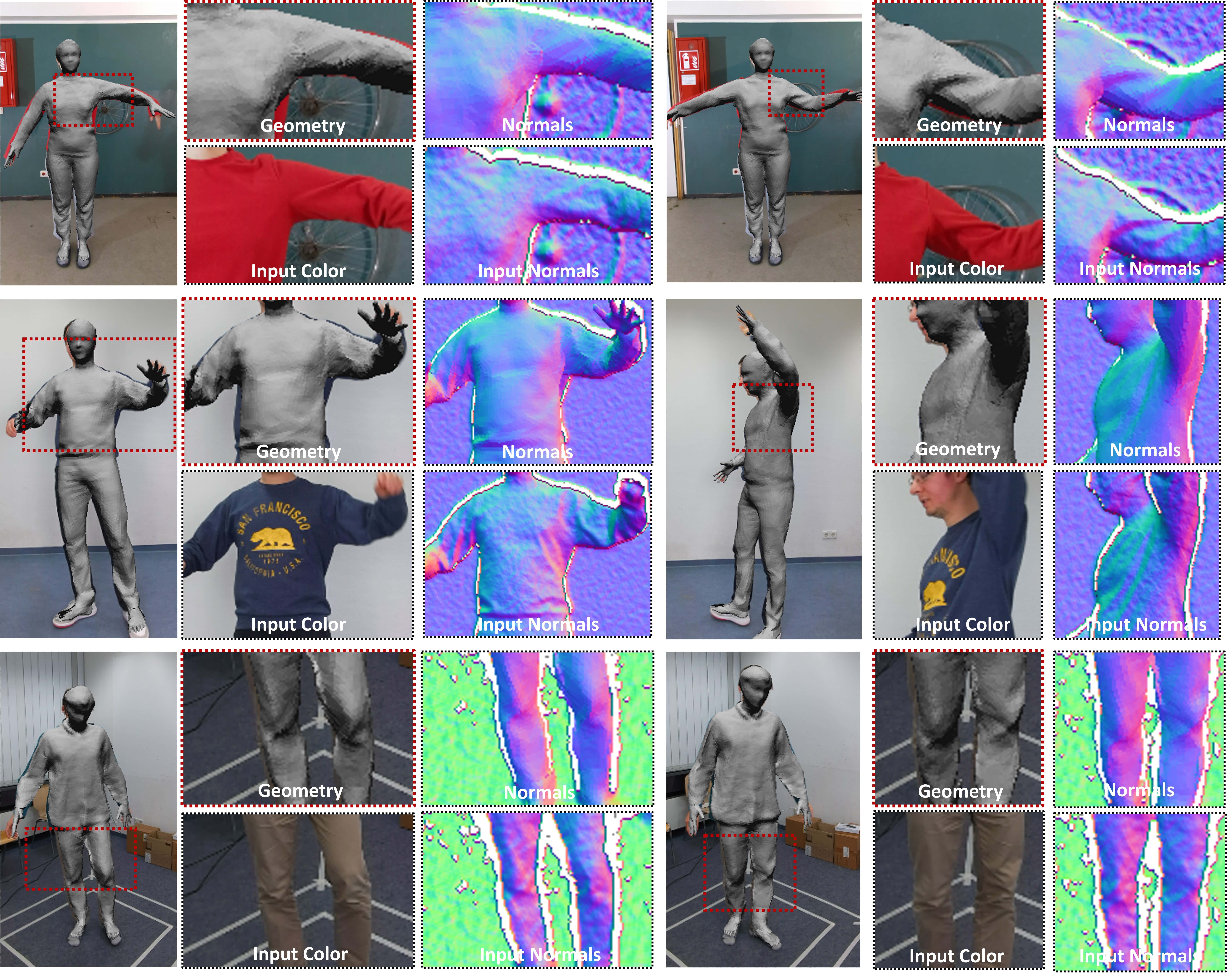}
    \vspace{-0.5cm}
    \caption{
        Dynamic Surface Function Networks are able to represent pose dependent wrinkles.
        Here, we show some sequences with corresponding close-ups to regions where pose dependent wrinkles occur (arms, upper-body and legs).
    }
    \label{fig:additional_closups}
\end{figure*}